\documentclass{article}

 \usepackage[final]{nips_2018}

\usepackage[utf8]{inputenc} 
\usepackage[T1]{fontenc}    
\usepackage{hyperref}       
\usepackage{url}            
\usepackage{booktabs}       
\usepackage{amsfonts,graphicx}       
\usepackage{amsmath} 
\usepackage{nicefrac}    
\usepackage{microtype}      
\usepackage{array}
\usepackage{subcaption}
\usepackage[dvipsnames]{xcolor}
\usepackage{authblk}

\title{Adversarial Machine Learning And Speech Emotion Recognition: Utilizing Generative Adversarial Networks For Robustness}

\author[1,2]{Siddique Latif}
\author[2]{Rajib Rana}
\author[1]{Junaid Qadir}

\affil[1]{Information Technology University (ITU)-Punjab, Pakistan}
\affil[2]{University of Southern Queensland, Australia}

\begin{document}

\maketitle

\begin{abstract}
  Deep learning has undoubtedly offered tremendous improvements in the performance of state-of-the-art speech emotion recognition (SER) systems. However, recent research on adversarial examples poses enormous challenges on the robustness of SER systems by showing the susceptibility of deep neural networks to adversarial examples as they rely only on small and imperceptible perturbations. In this study, we evaluate how adversarial examples can be used to attack SER systems and propose the first black-box adversarial attack on SER systems. We also explore potential defenses including adversarial training and generative adversarial network (GAN) to enhance robustness. Experimental evaluations suggest various interesting aspects of the effective utilization of adversarial examples useful for achieving robustness for SER systems opening up opportunities for researchers to further innovate in this space.
\end{abstract}

\section{Introduction}

Recent progress in machine learning (ML) is reinventing the future of intelligent systems enabling plethora of speech controlled applications \cite{cambria2016affective,poria2015towards,rana2016poster}. In particular, the emotion-aware systems are on the rise. The breakthrough in deep learning is largely fueling the development of highly accurate and robust emotion recognition systems  \cite{latif2018transfer,Latif2018}. 

Despite the superior performance of deep neural networks (DNNs), recent studies demonstrate that DNNs are highly vulnerable to the malicious attacks that use \textit{adversarial examples}. Adversarial examples are developed by malicious adversaries through the addition of unperceived perturbation with the intention of eliciting wrong responses from ML models. These adversarial examples can debilitate the performance of image recognition, object detection, and speech recognition models \cite{carlini2018audio}. Adversarial attacks can also be used to undermine the performance of speech-based emotion recognition (SER) systems \cite{gong2017crafting}, putting security-sensitive paralinguistic applications of SER systems at high risk. 

In this paper, we aim to investigate the utility of adversarial examples to achieve robustness in speech emotion classification to adversarial attacks. We consider a ``black-box'' attack that directly perturbs speech utterances with small and imperceptible noises. The generated adversarial examples are utilized within different schemes highlighting different aspects of robustness of SER systems. We further propose a GAN-based defense for SER systems and show that it can better resits adversarial examples compared to the previously proposed defense solutions such as adversarial training and random noise addition.  

\section{Background Literature and Motivation}
\label{gen_inst}

Existing methods of adversarial attacks including fast gradient sign method (FGSM) \cite{goodfellow6572explaining}, Jacobian-based saliency map attack (JSMA) \cite{papernot2016limitations}, DeepFool \cite{moosavi2016deepfool}, and Carlini and Wagner attacks \cite{carlini2017towards} compute the perturbation noise based on the gradient of targeted output with respect to the input. This is computed using backpropagation with the implicit assumption that the attacker has complete knowledge of the network and its parameters (such methods are called \textit{white-box} attacks). While the backpropagation method, which needs to compute the derivative of each layer of the network with respect to the input layers, can be efficiently applied in image recognition due to the  differentiability of all layers, the application of such methods is difficult for SER systems since these systems rely on complex acoustic features of the input audio utterances---such as Mel Frequency Cepstral Coefficients (MFCCs), spectrogram, extended Geneva Minimalistic Acoustic Parameter Set (eGeMAPS) \cite{eyben2016geneva}. The SER system's first layer is the pre-processing or the feature extraction layer, which does not offer an  efficient way to compute derivative, therefore, gradient-based methods \cite{papernot2016limitations,moosavi2016deepfool,carlini2017towards,su2017one} are not directly applicable to SER systems.

Adversarial attacks on ML have provoked an active area of research that is focusing on understanding the adversarial attack phenomenon \cite{fawzi2016robustness} and on techniques that can make ML models robust \cite{cisse2017parseval}. For speech-based systems, Carlini \cite{carlini2018audio} proposed a white-box iterative optimization-based attack for DeepSpeech \cite{hannun2014deep}, a state-of-the-art speech-to-text model, with 100\% success rate. Alzantot et al. \cite{alzantot2018did} proposed an adversarial attack on speech commands classification model by adding a small random noise (background noise) to the audio files. They achieved 87\% success without having any information of the underlying model. Song et al. \cite{song2017inaudible} proposed a mechanism that directly attacks the microphone used for sensing voice data and showed that an adversary can exploit the microphone’s non-linearity to control the targeted device with inaudible voice commands. Gong et al. \cite{gong2017crafting} presented an architecture to craft adversarial examples for computational paralinguistic applications. They perturbed the raw audio file and were able to cause a significant reduction in performance. Various other studies \cite{roy2017backdoor,iter2017generating,schonherr2018adversarial} have also developed adversarial attacks for speech recognition system. However, most of the previous research on targeted attacks for speech-based applications \cite{carlini2018audio,gong2017crafting,alzantot2018did,song2017inaudible,roy2017backdoor,iter2017generating,schonherr2018adversarial} has considered attacks on the model without investigating how adversarial examples may be utilized to make the ML models more robust. Our work is different since we not only propose an adversarial attack for SER system using adversarial examples but also leverage adversarial examples for making ML models more robust. 

\section{Proposed Audio Adversarial Examples}
\label{headings}

In this work, we adopt a simple approach to prepare adversarial examples by adding imperceptible noise ($\delta$) to the legitimate samples. We take an audio utterance $x$ with label $y$, and generate an adversarial example $x^{'}=x+\delta$ such that the SER system fails to correctly classify the given input while ensuring that $x$ and $x^{'}$ are very similar when perceived by humans. Previous speech-related studies have mostly considered ``non-real world'' random noise as adversarial noise. DolphinAttack exploits inaudible ultrasounds as adversarial noise to control the victim device inconspicuously but the attack sound was out of the human perception. Similarly, Alzantot et al. \cite{alzantot2018did} used random noise for creating an adversarial attack on speech recognition. It was however observed that the state-of-the-art classifiers are relatively robust to random noise \cite{fawzi2016robustness}. We therefore propose a black-box attack for SER system where an adversary can add ``real-world'' noise as adversarial perturbation. We empirically show that speech samples imputed with real-world noise can fool the classier while not being perceptible to the human ear. 

\textbf{Generation of $\delta$:} 

We use three noises: caf\'{e}, meeting, and station from the Demand Noise database \cite{thiemann2013diverse} and their imputation level is based on the already existing background noises (microphone noise and discussion noise) in the utterances. We estimate the existing noise in utterances using a well-known technique proposed in \cite{ephraim1984speech} that estimate noise using spectral and log-amplitude. We make the mean and variance of the above three noises equal to that of the reference noise. We also use $\epsilon$ as the variation parameter to further control the extent of perturbation and added the perturbation noise $\delta$ to the utterances using ($x_{i}+\epsilon\times\delta_{i}$). Where $x_{i}$ is $i^{th}$ utterance and $\delta_{i}$ is generated noise for it. In this way, the adversarial noise has a very small value similar to the existing noise and the adversarial example is unrecognizable to the human ear in the human perception test. 
Because this noise acts as the background noise it does not change the emotional context of a given audio file.   

\textbf{Human Perception and Classifier Test:} In order to assess the effect of added adversarial noise on the human listener, we asked five adults (age: 23-30 years) listeners to listen to 200 adversarial examples for different perturbation factor ($\epsilon$) and differentiate it from the original audio file. For the IEMOCAP and FAU-AIBO datasets, 96\% and 91\% of the samples were indistinguishable from the original utterances. When these examples were given to the classifier, the attack success rate was 72\% and 79\% for IEMOCAP and FAU-AIBO, respectively. 

\section{Experimental Setup and Results}
\label{Model}

We evaluated the generated adversarial examples using two well-known emotional corpora: IEMOCAP and FAU-AIBO. We consider binary classification problem (Positive and Negative) by mapping emotion to binary valance classes as used in \cite{latif2018transfer} and \cite{schuller2009interspeech}. Table \ref{Table:map} shows the considered emotions and their binary class mapping for both these datasets. We use the eGeMAPS features, a popular features set specifically suited for paralinguistic applications, for representing the audio samples. 

\begin{table}
  \caption{Binary class mapping of different emotions}
  \label{Table:map}
  \centering
  \begin{tabular}{lll}
    \toprule
    Dataset & Positive class     & Negative Class \\
    \midrule
    IEMOCAP & happiness, exited, neutral  & anger, sadness     \\
    FAU-AIBO     & neutral, motherese, and joyful & angry, touchy, reprimanding, and emphatic \\
    \bottomrule
  \end{tabular}
\end{table}

\textbf{Classification Model:} We consider LSTM-RNN for emotion classification. LSTM is a popular RNN and widely employed in audio \cite{latif2018phonocardiographic} and emotion classification \cite{Latif2018} due to their ability to model contextual information. We find the best model structure by evaluating different number of layers. We obtained the best results with two LSTM layers, one dense layer, and a softmax as the last layer. We initially used a learning rate of 0.002 to start training the model and halved this rate after every 5 epochs if performance did not improve on the test set. This process stopped when the learning rate reached below 0.00001.

\begin{figure}[!ht]%
\centering
\begin{subfigure}{0.5\linewidth}
\includegraphics[trim=0.1cm 0cm 0cm 0.1cm,clip=true,width=\linewidth]{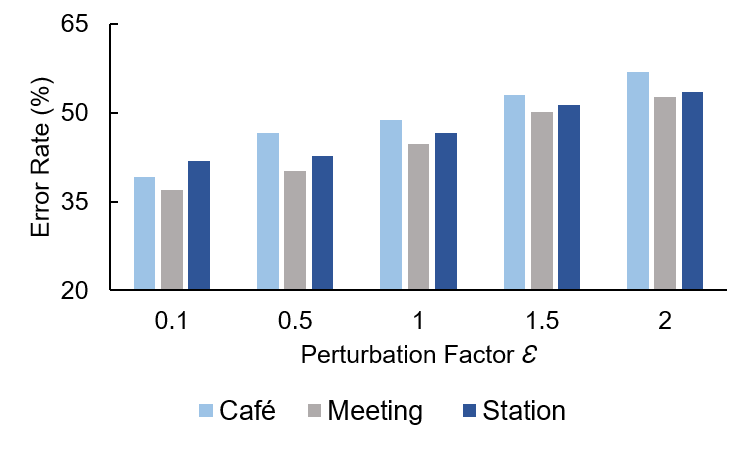}%
\captionsetup{justification=centering}
\caption{}%
\label{AD_FAU}%
\end{subfigure}
\begin{subfigure}{0.5\linewidth}
\includegraphics[trim=0.1cm 0cm 0cm 0.1cm,clip=true,width=\linewidth]{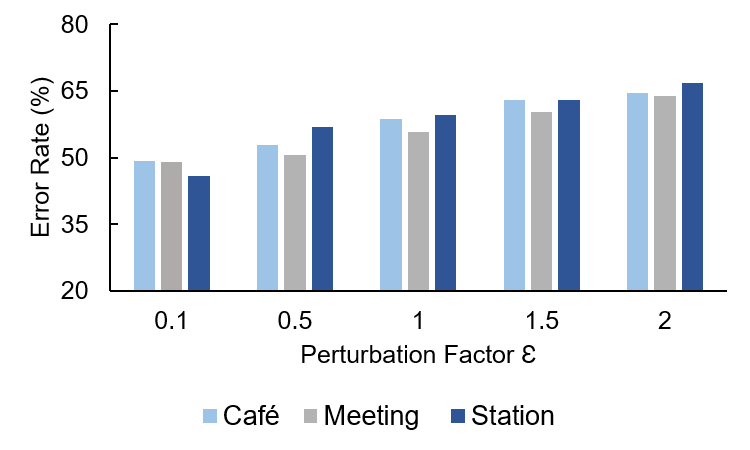}%
\captionsetup{justification=centering}
\caption{} %
\label{AD_IEMOCAP}%
\end{subfigure}%
\caption{The error rate (\%) with different perturbation factors for speech emotion classification for FAU-AIBO (left) and IEMOCAP (right) datasets.}
\label{fig:Adversarial}
\end{figure}

\textbf{Emotion Classification Results:} For experimentation, we evaluated the model in a  speaker independent scheme. IEMOCAP dataset consists of five sessions; we used four session for training and one for testing, consistent with the methodology of previous studies \cite{latif2018transfer, Latif2018}. For FAU-AIBO, we followed the speaker-independent training strategy proposed in the 2009 Interspeech Emotion Challenge \cite{schuller2009interspeech}. For emotion classification on legitimate examples, we achieved $68.35\%$ and $56.41\%$ unweighted accuracy (UA) on FAU-AIBO and IEMOCAP dataset, respectively. The results on adversarial examples are compared with these results. We generated adversarial examples with different values of $\epsilon$ (0.1--2) to evaluate the performance of model with different perturbation factor. This is demonstarted in Figure \ref{fig:Adversarial} presents the emotion classification error on adversarial samples with different values of $\epsilon$. 

Based on Figure \ref{fig:Adversarial} the proposed attack is effective in fooling the classifier for emotion classification tasks. With the perturbation factor 2.0, the classification error rate is increased from 31.65 and 43.59
to 56.87 and 66.87 for FAU-AIBO and IEMOCAP dataset respectively.

\section{Defense Mechanisms}
\subsection{Training with Adversarial Examples}
\begin{figure}[!ht]%
\centering
\begin{subfigure}{0.5\linewidth}
\includegraphics[trim=0.1cm 0cm 0cm 0.1cm,clip=true,width=\linewidth]{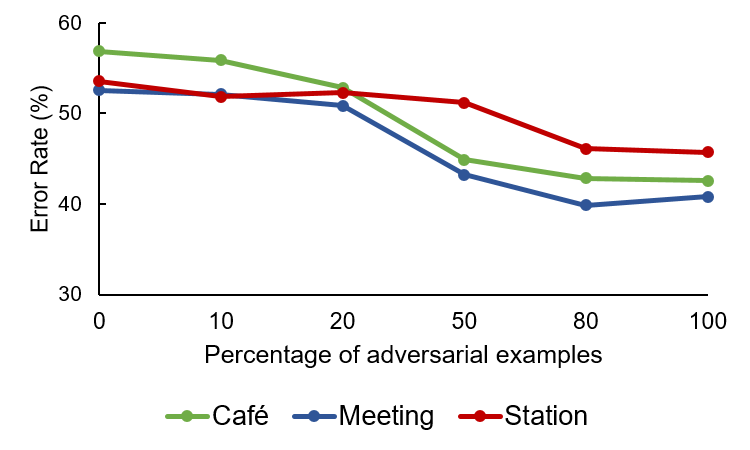}%
\captionsetup{justification=centering}
\caption{}%
\label{AD_AIBO_TR}%
\end{subfigure}
\begin{subfigure}{0.5\linewidth}
\includegraphics[trim=0.1cm 0cm 0cm 0.1cm,clip=true,width=\linewidth]{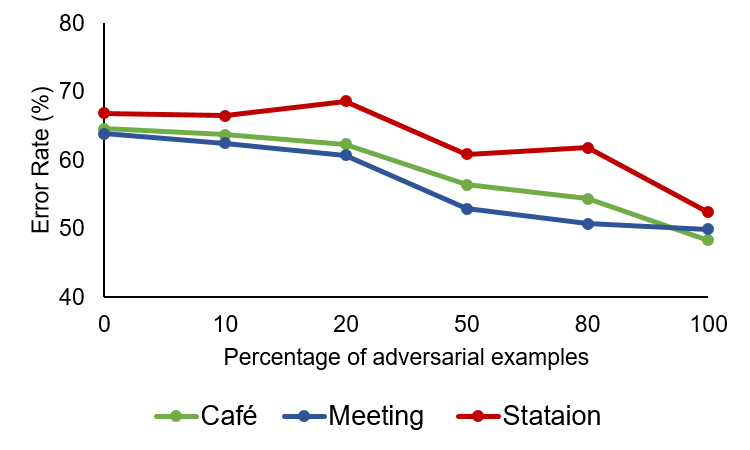}%
\captionsetup{justification=centering}
\caption{} %
\label{AD_IEMOCAP_TR}%
\end{subfigure}%
\caption{The error rate (\%) with varying the percentage of adversarial samples as training data for FAU-AIBO (left) and IEMOCAP (right) datasets.}
\label{fig:Ad_TR}
\end{figure}

Adversarial training of model is considered as a possible defense to adversarial attacks when the exact nature of the attack is known. Model training on the mixture of clean and adversarial examples can somewhat help regularization \cite{szegedy2013intriguing}. Training on adversarial samples is different from data augmentation methods that are performed based on the expected translations in test data. To the best of our knowledge, adversarial training is not explored for SER systems and other speech/audio classification systems. We explore this phenomenon by mixing adversarial examples with training data to highlight the robustness of model against attack. We trained the model with training data comprising of a varying percentage of adversarial examples (10\% to 100\% of training data). Figure \ref{fig:Ad_TR} shows the classification error rate (\%)  significantly decreases with the increase of percentage of adversarial examples in the training data. 

\subsection{Training with Random Noise}
It is reported in \cite{liu2017towards} that the addition of a random noise layer to the neural network can prevent strong gradient-based attacks in the image domain. We evaluated this phenomenon in speech emotion classification system by adding a small random noise to overall training data and evaluated the performance against the proposed attacks. Table \ref{table:random} shows that emotion classification error reduces only slightly with the addition of random noise in training data, which indicates that this strategy is not particularly effective in the SER settings. 

\begin{table}[!ht]
\caption{Emotion classification error (\%) while adding random noise in training data}
\centering
\scriptsize
\captionsetup{width=0.9\textwidth}
\begin{tabular}{|m{1.2cm}|m{2.6cm}|m{3.5cm}|m{3.5cm}|}
\hline
Dataset & Adversarial Perturbations & Error (max) with adversarial attack & Error by training with random noise 
\\ \hline
&\begin{tabular}[c]{@{}l@{}}Caf\'{e} \end{tabular}
&\begin{tabular}[c]{@{}l@{}}56.87\end{tabular}
&\begin{tabular}[c]{@{}l@{}}54.02\end{tabular}
\\\cline{2-4}
\begin{tabular}[c]{@{}l@{}}FAU-AIBO \end{tabular}
&\begin{tabular}[c]{@{}l@{}}Meeting\end{tabular}
&\begin{tabular}[c]{@{}l@{}}52.58\end{tabular}
&\begin{tabular}[c]{@{}l@{}}49.24\end{tabular}
\\\cline{2-4}
&\begin{tabular}[c]{@{}l@{}}Station\end{tabular}
&\begin{tabular}[c]{@{}l@{}}53.57\end{tabular}
&\begin{tabular}[c]{@{}l@{}}48.51\end{tabular}
\\\hline
&\begin{tabular}[c]{@{}l@{}}Caf\'{e} \end{tabular}
&\begin{tabular}[c]{@{}l@{}}64.58\end{tabular}
&\begin{tabular}[c]{@{}l@{}}56.73\end{tabular}
\\\cline{2-4}
\begin{tabular}[c]{@{}l@{}}IEMOCAP\end{tabular}
&\begin{tabular}[c]{@{}l@{}}Meeting \end{tabular}
&\begin{tabular}[c]{@{}l@{}}63.88\end{tabular}
&\begin{tabular}[c]{@{}l@{}}52.57\end{tabular}
\\\cline{2-4}
&\begin{tabular}[c]{@{}l@{}}Station \end{tabular}
&\begin{tabular}[c]{@{}l@{}}66.87\end{tabular}
&\begin{tabular}[c]{@{}l@{}}60.87\end{tabular}
\\\hline
\end{tabular}
\vspace{.5mm}
\label{table:random}
\end{table}

\subsection{Using Generative Adversarial Network}

Generative adversarial networks (GANs) \cite{goodfellow2014generative} are deep models that learn to generate samples, ideally indistinguishable
from the real data $x$, that are supposed to belong to an unknown data distribution, $p_{data}(x)$. GANs consist of two networks, a generator ($G$) and a discriminator ($D$). The generator network ($G$) maps latent vectors from some known prior $p_z$ to samples and discriminator tasked to differentiate between the real sample $x$ or fake $G(z)$. Mathematically, this is represented by the following optimization program:
\begin{equation}
    \underset{G}{\text{min}} \  \underset{D}{\text{max}}
    \quad
    \mathrm{E}_x[\log(D(x))] + \mathrm{E}_y[\log(1 - D(G(z)))]
\end{equation}
where $G$ and $D$ play this game to fool each other using this min-max optimization program. 
In our case, $G$ network is tasked to remove the adversarial noise from the adversarial examples $z$. The $G$ network is structured like an autoencoder using LSTM layers. 
In the $G$ network, the encoder part compresses the contextual (emotional) information of the input speech features and the decoder uses this representation for reconstruction. The $D$ network follows the same encoder-decoder architecture. For training $G$ and $D$ for different possible scenarios, we used the training data from both the datasets to train the GAN. For each $G$ step, the discriminator was updated twice. For faster convergence, we pretrained the $G$ network in each case. We trained GAN using RMSProp optimizer with learning rate $1 \times 10^{-4}$ and batch size of $32$, until convergence. For training we used utterances corrupted by the three adversarial noises: caf\'{e}, meeting, station as noisy data and it was tasked to clean the utterances. Data cleaned by GAN was given to the classifier for emotion classification. 

Table \ref{table:gan} shows emotion classification results on audio utterances cleaned by GAN. It can be noted that the classification error significantly reduces by removing adversarial noise from data prior to classification.

\begin{table}[!ht]
\caption{Emotion classification error (\%) by utilizing GAN as defense against adversarial noise removal}
\centering
\scriptsize
\captionsetup{width=0.9\textwidth}
\begin{tabular}{|m{1.2cm}|m{2.8cm}|m{3.5cm}|m{3cm}|}
\hline
Dataset & Adversarial Perturbations & Error (max) with adversarial attack & Error by employing GAN prior to classification 
\\ \hline
&\begin{tabular}[c]{@{}l@{}}Caf\'{e} \end{tabular}
&\begin{tabular}[c]{@{}l@{}}68.82\end{tabular}
&\begin{tabular}[c]{@{}l@{}}38.31\end{tabular}
\\\cline{2-4}
\begin{tabular}[c]{@{}l@{}}FAU-AIBO \end{tabular}
&\begin{tabular}[c]{@{}l@{}}Meeting\end{tabular}
&\begin{tabular}[c]{@{}l@{}}62.58\end{tabular}
&\begin{tabular}[c]{@{}l@{}}36.02\end{tabular}
\\\cline{2-4}
&\begin{tabular}[c]{@{}l@{}}Station\end{tabular}
&\begin{tabular}[c]{@{}l@{}}66.87\end{tabular}
&\begin{tabular}[c]{@{}l@{}}35.14\end{tabular}
\\\hline
&\begin{tabular}[c]{@{}l@{}}Caf\'{e} \end{tabular}
&\begin{tabular}[c]{@{}l@{}}65.87\end{tabular}
&\begin{tabular}[c]{@{}l@{}}49.20\end{tabular}
\\\cline{2-4}
\begin{tabular}[c]{@{}l@{}}IEMOCAP\end{tabular}
&\begin{tabular}[c]{@{}l@{}}Meeting \end{tabular}
&\begin{tabular}[c]{@{}l@{}}67.70\end{tabular}
&\begin{tabular}[c]{@{}l@{}}48.18\end{tabular}
\\\cline{2-4}
&\begin{tabular}[c]{@{}l@{}}Station \end{tabular}
&\begin{tabular}[c]{@{}l@{}}69.87\end{tabular}
&\begin{tabular}[c]{@{}l@{}}46.24\end{tabular}
\\\hline
\end{tabular}
\vspace{.5mm}
\label{table:gan}
\end{table}

\section{Discussion}

\begin{figure}[!ht]%
\centering
\begin{subfigure}{0.5\linewidth}
\includegraphics[trim=0.1cm 0cm 0cm 0.1cm,clip=true,width=\linewidth]{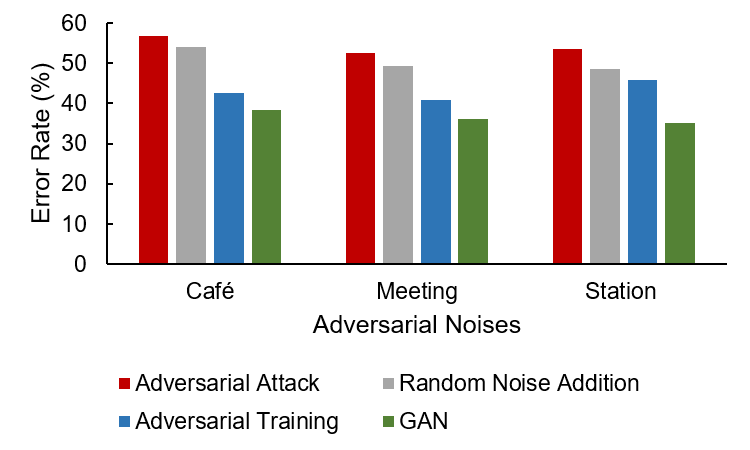}%
\captionsetup{justification=centering}
\caption{}%
\label{AD_FAU_a}%
\end{subfigure}
\begin{subfigure}{0.5\linewidth}
\includegraphics[trim=0.1cm 0cm 0cm 0.1cm,clip=true,width=\linewidth]{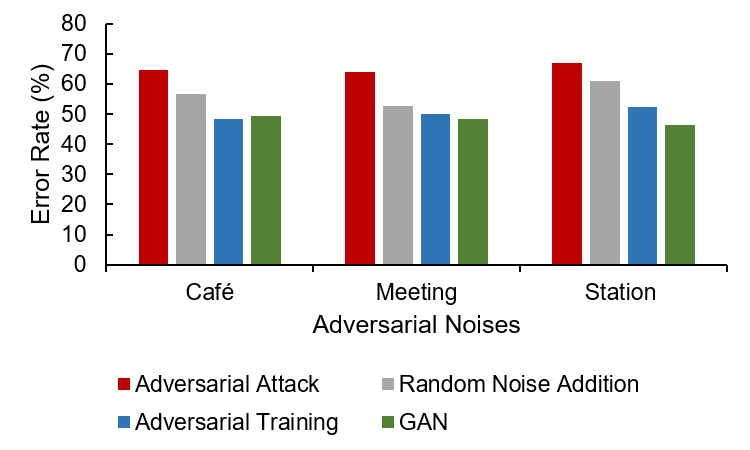}%
\captionsetup{justification=centering}
\caption{} %
\label{AD_IEMOCAP_b}%
\end{subfigure}%
\caption{The error rate (\%) with three different approaches against adversarial examples for FAU-AIBO (left) and IEMOCAP (right) datasets.}
\label{fig:diss}
\end{figure}

From the experimental evaluations we find that GAN-based defense against adversarial audio examples better withstands adversarial examples compared to other approaches. Figure \ref{fig:diss} shows a comparison of the different defense mechanisms using two well-known datasets: IEMOCAP and FAU-AIBO. The addition of random noise in training utterances slightly reduces speech emotion classification error, however, using adversarial training, classification error is significantly reduced. This supports that training with random noise is not adequate to avoid adversarial attacks.  
The best results are however achieved using GAN. 
This motivates further research for its utilization in other speech-based intelligent systems for the minimization of adversarial perturbations. It is worth pointing out that GANs require information about the exact type and nature of adversarial examples for its training, but this is also an essential requirement for the adversarial training mechanism.

\section{Conclusions}

In this paper, we propose a black-box method to generate adversarial perturbations in audio examples of speech emotion recognition system (SER). We also propose a defence strategy using Generative Adversarial Network (GAN) for enhancing the robustness of SER system by first cleaning the perturbed utterances through GANs and then running a classifier on it. We compared our GAN-based defense against adversarial training and the addition of random noise in training examples and showed that our GAN-based defense provides consistently better results in speech emotion recognition. We anticipate that the attack and defense that we propose can also be utilized more generally for other speech-based intelligent systems. 


\begin{thebibliography}{10}

\bibitem{cambria2016affective}
Erik Cambria.
\newblock Affective computing and sentiment analysis.
\newblock {\em IEEE Intelligent Systems}, 31(2):102--107, 2016.

\bibitem{poria2015towards}
Soujanya Poria, Erik Cambria, Amir Hussain, and Guang-Bin Huang.
\newblock Towards an intelligent framework for multimodal affective data
  analysis.
\newblock {\em Neural Networks}, 63:104--116, 2015.

\bibitem{rana2016poster}
Rajib Rana.
\newblock Poster: Context-driven mood mining.
\newblock In {\em Proceedings of the 14th Annual International Conference on
  Mobile Systems, Applications, and Services Companion}, pages 143--143. ACM,
  2016.

\bibitem{latif2018transfer}
Siddique Latif, Rajib Rana, Shahzad Younis, Junaid Qadir, and Julien Epps.
\newblock Transfer learning for improving speech emotion classification
  accuracy.
\newblock {\em Proc. Interspeech 2018}, pages 257--261, 2018.

\bibitem{Latif2018}
Siddique Latif, Rajib Rana, Junaid Qadir, and Julien Epps.
\newblock Variational autoencoders for learning latent representations of
  speech emotion: A preliminary study.
\newblock In {\em Proc. Interspeech 2018}, pages 3107--3111, 2018.

\bibitem{carlini2018audio}
Nicholas Carlini and David Wagner.
\newblock Audio adversarial examples: Targeted attacks on speech-to-text.
\newblock {\em arXiv preprint arXiv:1801.01944}, 2018.

\bibitem{gong2017crafting}
Yuan Gong and Christian Poellabauer.
\newblock Crafting adversarial examples for speech paralinguistics
  applications.
\newblock {\em arXiv preprint arXiv:1711.03280}, 2017.

\bibitem{goodfellow6572explaining}
Ian~J Goodfellow, Jonathon Shlens, and Christian Szegedy.
\newblock Explaining and harnessing adversarial examples (2014).
\newblock {\em arXiv preprint arXiv:1412.6572}.

\bibitem{papernot2016limitations}
Nicolas Papernot, Patrick McDaniel, Somesh Jha, Matt Fredrikson, Z~Berkay
  Celik, and Ananthram Swami.
\newblock The limitations of deep learning in adversarial settings.
\newblock In {\em Security and Privacy (EuroS\&P), 2016 IEEE European Symposium
  on}, pages 372--387. IEEE, 2016.

\bibitem{moosavi2016deepfool}
Seyed-Mohsen Moosavi-Dezfooli, Alhussein Fawzi, and Pascal Frossard.
\newblock Deepfool: a simple and accurate method to fool deep neural networks.
\newblock In {\em Proceedings of the IEEE Conference on Computer Vision and
  Pattern Recognition}, pages 2574--2582, 2016.

\bibitem{carlini2017towards}
Nicholas Carlini and David Wagner.
\newblock Towards evaluating the robustness of neural networks.
\newblock In {\em 2017 IEEE Symposium on Security and Privacy (SP)}, pages
  39--57. IEEE, 2017.

\bibitem{eyben2016geneva}
Florian Eyben, Klaus~R Scherer, Bj{\"o}rn~W Schuller, Johan Sundberg, Elisabeth
  Andr{\'e}, Carlos Busso, Laurence~Y Devillers, Julien Epps, Petri Laukka,
  Shrikanth~S Narayanan, et~al.
\newblock The geneva minimalistic acoustic parameter set (gemaps) for voice
  research and affective computing.
\newblock {\em IEEE Transactions on Affective Computing}, 7(2):190--202, 2016.

\bibitem{su2017one}
Jiawei Su, Danilo~Vasconcellos Vargas, and Sakurai Kouichi.
\newblock One pixel attack for fooling deep neural networks.
\newblock {\em arXiv preprint arXiv:1710.08864}, 2017.

\bibitem{fawzi2016robustness}
Alhussein Fawzi, Seyed-Mohsen Moosavi-Dezfooli, and Pascal Frossard.
\newblock Robustness of classifiers: from adversarial to random noise.
\newblock In {\em Advances in Neural Information Processing Systems}, pages
  1632--1640, 2016.

\bibitem{cisse2017parseval}
Moustapha Cisse, Piotr Bojanowski, Edouard Grave, Yann Dauphin, and Nicolas
  Usunier.
\newblock Parseval networks: Improving robustness to adversarial examples.
\newblock {\em arXiv preprint arXiv:1704.08847}, 2017.

\bibitem{hannun2014deep}
Awni Hannun, Carl Case, Jared Casper, Bryan Catanzaro, Greg Diamos, Erich
  Elsen, Ryan Prenger, Sanjeev Satheesh, Shubho Sengupta, Adam Coates, et~al.
\newblock Deep speech: Scaling up end-to-end speech recognition.
\newblock {\em arXiv preprint arXiv:1412.5567}, 2014.

\bibitem{alzantot2018did}
Moustafa Alzantot, Bharathan Balaji, and Mani Srivastava.
\newblock Did you hear that? adversarial examples against automatic speech
  recognition.
\newblock {\em arXiv preprint arXiv:1801.00554}, 2018.

\bibitem{song2017inaudible}
Liwei Song and Prateek Mittal.
\newblock Inaudible voice commands.
\newblock {\em arXiv preprint arXiv:1708.07238}, 2017.

\bibitem{roy2017backdoor}
Nirupam Roy, Haitham Hassanieh, and Romit Roy~Choudhury.
\newblock Backdoor: Making microphones hear inaudible sounds.
\newblock In {\em Proceedings of the 15th Annual International Conference on
  Mobile Systems, Applications, and Services}, pages 2--14. ACM, 2017.

\bibitem{iter2017generating}
Dan Iter, Jade Huang, and Mike Jermann.
\newblock Generating adversarial examples for speech recognition.
\newblock \url{http://web.stanford.edu/class/cs224s/reports/Dan_Iter.pdf},
  2017.

\bibitem{schonherr2018adversarial}
Lea Sch{\"o}nherr, Katharina Kohls, Steffen Zeiler, Thorsten Holz, and Dorothea
  Kolossa.
\newblock Adversarial attacks against automatic speech recognition systems via
  psychoacoustic hiding.
\newblock {\em arXiv preprint arXiv:1808.05665}, 2018.

\bibitem{thiemann2013diverse}
Joachim Thiemann, Nobutaka Ito, and Emmanuel Vincent.
\newblock The diverse environments multi-channel acoustic noise database: A
  database of multichannel environmental noise recordings.
\newblock {\em The Journal of the Acoustical Society of America},
  133(5):3591--3591, 2013.

\bibitem{ephraim1984speech}
Yariv Ephraim and David Malah.
\newblock Speech enhancement using a minimum-mean square error short-time
  spectral amplitude estimator.
\newblock {\em IEEE Transactions on acoustics, speech, and signal processing},
  32(6):1109--1121, 1984.

\bibitem{schuller2009interspeech}
Bj{\"o}rn Schuller, Stefan Steidl, and Anton Batliner.
\newblock The interspeech 2009 emotion challenge.
\newblock In {\em Tenth Annual Conference of the International Speech
  Communication Association}, 2009.

\bibitem{latif2018phonocardiographic}
Siddique Latif, Muhammad Usman, Rajib Rana, and Junaid Qadir.
\newblock Phonocardiographic sensing using deep learning for abnormal heartbeat
  detection.
\newblock {\em IEEE Sensors Journal}, 2018.

\bibitem{szegedy2013intriguing}
Christian Szegedy, Wojciech Zaremba, Ilya Sutskever, Joan Bruna, Dumitru Erhan,
  Ian Goodfellow, and Rob Fergus.
\newblock Intriguing properties of neural networks.
\newblock {\em arXiv preprint arXiv:1312.6199}, 2013.

\bibitem{liu2017towards}
Xuanqing Liu, Minhao Cheng, Huan Zhang, and Cho-Jui Hsieh.
\newblock Towards robust neural networks via random self-ensemble.
\newblock {\em arXiv preprint arXiv:1712.00673}, 2017.

\bibitem{goodfellow2014generative}
Ian Goodfellow, Jean Pouget-Abadie, Mehdi Mirza, Bing Xu, David Warde-Farley,
  Sherjil Ozair, Aaron Courville, and Yoshua Bengio.
\newblock Generative adversarial nets.
\newblock In {\em Advances in neural information processing systems}, pages
  2672--2680, 2014.

\end{thebibliography}

\end{document}